%% file: root.tex
\documentclass[letterpaper, 10 pt, conference]{ieeeconf}  % Comment this line out if you need a4paper

\IEEEoverridecommandlockouts                              % This command is only needed if 
\usepackage{graphics} % for pdf, bitmapped graphics files
\usepackage{epsfig} % for postscript graphics files
\usepackage{mathptmx} % assumes new font selection scheme installed
\usepackage{times} % assumes new font selection scheme installed
\usepackage{amsmath} % assumes amsmath package installed
\usepackage{amssymb}  % assumes amsmath package installed
\usepackage{hyperref}
\usepackage{booktabs}
\usepackage{array}
\usepackage{graphicx}
\usepackage{multirow}
\usepackage{tcolorbox}
\usepackage{listings}
\usepackage[table]{xcolor}
\usepackage{colortbl}
\title{\LARGE \bf
AdaClearGrasp: Learning Adaptive Clearing for Zero-Shot Robust Dexterous Grasping in Densely Cluttered Environments
}

\author{Zixuan Chen$^{1*}$, Wenquan Zhang$^{1*}$, Jing Fang$^{2}$, Ruiming Zeng$^{1}$, Zhixuan Xu$^{3}$, Yiwen Hou$^{3}$, Xinke Wang$^{1}$, \\ Jieqi Shi$^{1}$, Jing Huo$^{1}$, Yang Gao$^{1}$% <-this % stops a space
\thanks{*Equal Contribution}% <-this % stops a space
\thanks{$^{1}$Nanjing University, $^{2}$Nanjing University of Posts and Telecommunications, $^{3}$National University of Singapore}%
}

\begin{document}

\maketitle
\thispagestyle{empty}
\pagestyle{empty}

%%%%%%%%%%%%%%%%%%%%%%%%%%%%%%%%%%%%%%%%%%%%%%%%%%%%%%%%%%%%%%%%%%%%%%%%%%%%%%%%

\input{section/1_abs}
\input{section/2_intro}

\input{section/3_related_works}

\input{section/4_method}

\input{section/5_exp}

\input{section/6_conclusion}



% \section*{ACKNOWLEDGMENT}

% The preferred spelling of the word ÒacknowledgmentÓ in America is without an ÒeÓ after the ÒgÓ. Avoid the stilted expression, ÒOne of us (R. B. G.) thanks . . .Ó  Instead, try ÒR. B. G. thanksÓ. Put sponsor acknowledgments in the unnumbered footnote on the first page.

% %%%%%%%%%%%%%%%%%%%%%%%%%%%%%%%%%%%%%%%%%%%%%%%%%%%%%%%%%%%%%%%%%%%%%%%%%%%%%%%%

% References are important to the reader; therefore, each citation must be complete and correct. If at all possible, references should be commonly available publications.

{\small
\bibliographystyle{IEEEtran}
\bibliography{reference}
}

\clearpage
\input{section/7_appendix}

\end{document}

%% file: section/1_abs.tex
%%%%%%%%%%%%%%%%%%%%%%%%%%%%%%%%%%%%%%%%%%%%%%%%%%%%%%%%%%%%%%%%%%%%%%%%%%%%%%%%
\begin{abstract}
In densely cluttered environments, physical interference, visual occlusions, and unstable contacts often cause direct dexterous grasping to fail, while aggressive singulation strategies may compromise safety. Enabling robots to adaptively decide whether to clear surrounding objects or directly grasp the target is therefore crucial for robust manipulation.
We propose AdaClearGrasp, a closed-loop decision–execution framework for adaptive clearing and zero-shot dexterous grasping in densely cluttered environments. The framework formulates manipulation as a controllable high-level decision process that determines whether to directly grasp the target or first clear surrounding objects. A pretrained vision–language model (VLM) interprets visual observations and language task descriptions to reason about grasp interference and generate a high-level planning skeleton, which invokes structured atomic skills through a unified action interface. For dexterous grasping, we train a reinforcement learning policy with a relative hand–object distance representation, enabling zero-shot generalization across diverse object geometries and physical properties. During execution, visual feedback monitors outcomes and triggers replanning upon failures, forming a closed-loop correction mechanism.
To evaluate language-conditioned dexterous grasping in clutter, we introduce \texttt{Clutter-Bench}, the first simulation benchmark with graded clutter complexity. It includes seven target objects across three clutter levels, yielding 210 task scenarios. We further perform sim-to-real experiments on three objects under three clutter levels (18 scenarios). Results demonstrate that AdaClearGrasp significantly improves grasp success rates in densely cluttered environments.
For more videos and code, please visit our project website: \href{https://chenzixuan99.github.io/adaclear-grasp.github.io/}{https://chenzixuan99.github.io/adaclear-grasp.github.io/}.
\end{abstract}

%% file: section/2_intro.tex
\section{INTRODUCTION}

Deploying robotic manipulation systems in real-world environments requires robust dexterous grasping in densely cluttered scenes~\cite{IHM_survey,chen2022system,il_hand_craft}. For instance, in kitchen organization tasks, a robot may need to retrieve a target object from tightly packed cups, plates, and utensils. In such environments, target objects are often surrounded by other items, causing physical interference, visual occlusions, and unstable contacts that make direct grasping unreliable. Prior work shows that effective manipulation in clutter often requires interacting with surrounding objects, such as pushing, singulating, or rearranging them, to expose the target~\cite{dogar2012planning,zeng2018learning,morrison2018closing,xu2025dexsingrasp}. While removing nearby objects can mitigate these issues, excessive clearing may introduce unnecessary interactions and increase the risk of damaging surrounding items. Therefore, robust dexterous manipulation in clutter requires not only precise low-level control but also high-level decision-making to determine \emph{when} and \emph{how} to perform clearing before grasping.

\begin{figure}[t!]
    \centering
    \includegraphics[width=\linewidth]{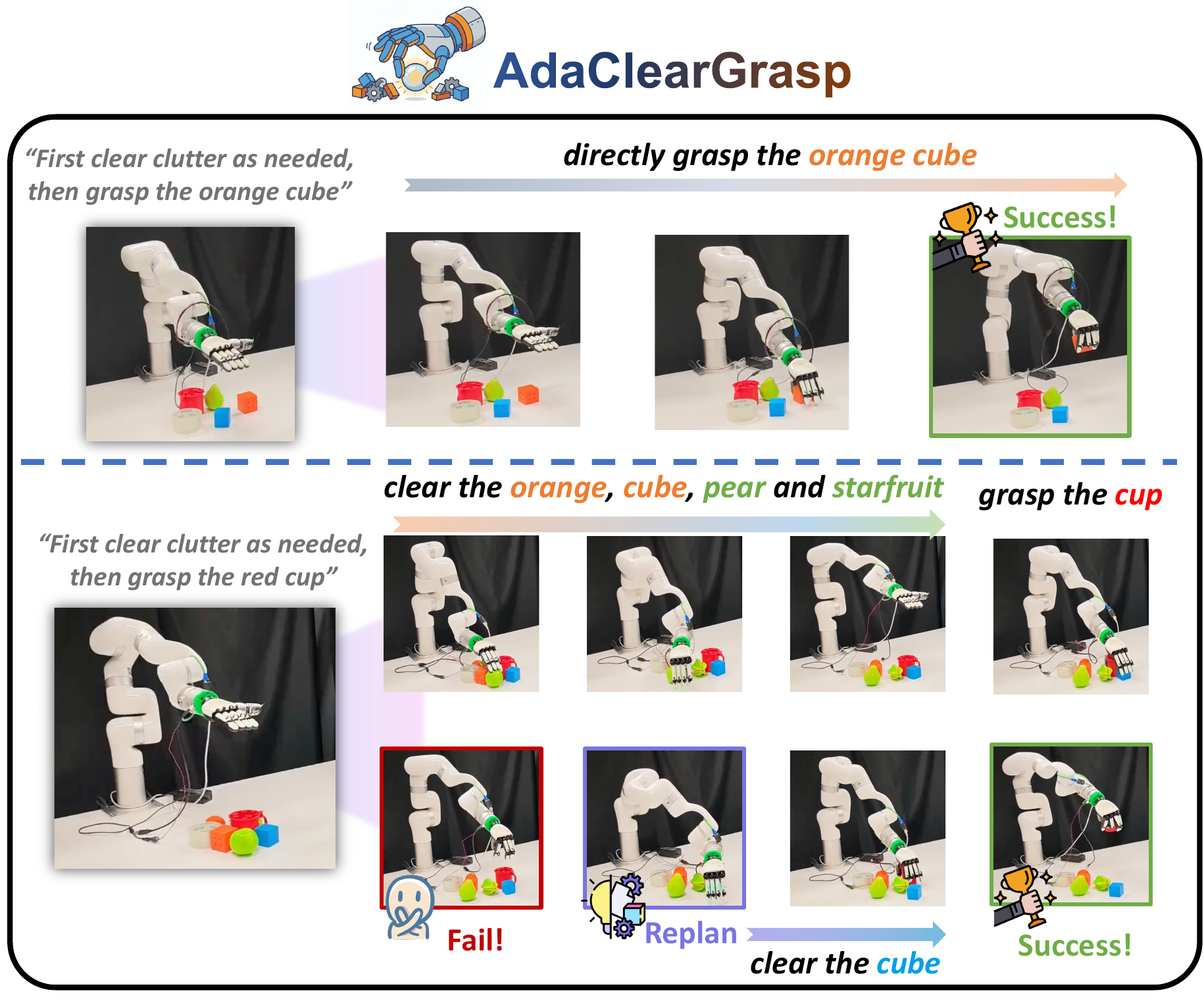}
    \caption{We propose \textbf{AdaClearGrasp}, a closed-loop system that leverages VLM-based reasoning to perform adaptive clearing for robust dexterous grasping of target objects in densely cluttered scenes.}
    \label{fig:teaser}
    \vspace{-6mm}
\end{figure}

Most existing dexterous grasping methods assume relatively isolated targets or rely on end-to-end reinforcement learning (RL) policies that map perception directly to control. Recent work has explored unified representations for dexterous grasping and cross-embodiment manipulation~\cite{openai2019dexterous,wei2024drograsp}. ClutterDexGrasp~\cite{chen2025clutterdexgrasp} further extends RL to cluttered scenes using a teacher–student framework with clutter-density curricula, enabling implicit behaviors such as obstacle nudging.
While effective for general grasping, these methods largely treat manipulation as a low-level control problem and rely on emergent strategies that struggle in dense clutter where long-horizon reasoning and adaptive interactions are required. In addition, pure RL approaches are often sample-inefficient and sensitive to scene variations.

Recent advances in vision-language models (VLMs) have demonstrated strong capabilities in semantic understanding and high-level reasoning for robotic manipulation~\cite{chen2026deco,chengravmad,lynch2020language,ahn2022saycan,driess2023palm,brohan2023rt2}. VLMs have also been applied to guide low-level dexterous control. For example, \cite{de2025scaffolding} provides coarse task-level guidance, while \cite{liu2025robodexvlm} generates executable actions and replans after failures.
However, these approaches have limitations. The former handles only a single object and uses open-loop execution, while the latter offers limited feedback and is not designed for densely cluttered scenes. In such scenarios, deciding whether and how to clear surrounding objects is a multi-constraint problem involving visibility, kinematic reachability, and interaction safety, which must be tightly coupled with low-level control. Open-loop reasoning or simple task-level replanning often fails to satisfy these constraints reliably.

Therefore, a practical system in densely cluttered environments must continuously incorporate environmental feedback and adapt strategies during execution to ensure robust performance. To this end, we propose \textbf{AdaClearGrasp}, a closed-loop framework that integrates hierarchical VLM reasoning with dexterous manipulation policies to enable adaptive clearing for zero-shot robust grasping.
As shown in Fig.~\ref{fig:teaser}, AdaClearGrasp uses a VLM-based planner to reason about graspability and occlusion by jointly analyzing visual observations and language instructions, determining whether surrounding objects obstruct the target grasp and whether clearing is required. When interference is detected, the planner generates high-level action sequences that invoke parameterized atomic clearing skills to rearrange nearby objects.
Once the target becomes accessible, the system executes a RL-based dexterous grasping policy, \textbf{GeoGrasp}. The policy leverages geometric relations between the hand and object, represented by nearest-neighbor vectors from hand links to the object point cloud. This geometry-aware representation enables robust dynamic grasping and facilitates zero-shot generalization to unseen object geometries and physical properties.
During execution, AdaClearGrasp maintains a closed-loop mechanism that monitors progress through visual feedback and triggers replanning or recovery upon failures. Through structured action interfaces and failure feedback, high-level decisions are continuously corrected, improving robustness under real-world contact uncertainties. This integration of semantic reasoning and dexterous control enables reliable long-horizon manipulation in cluttered environments.

To systematically evaluate language-conditioned dexterous grasping in clutter, we introduce \textbf{\texttt{Clutter-Bench}}, a graded benchmark built on ManiSkill3~\cite{mu2021maniskill}. The benchmark includes seven target object categories from simple geometric shapes to complex household items and defines three difficulty levels based on the number of obstacles (2, 4, and 6 objects). By providing natural language task specifications and reproducible scene configurations, \textbf{\texttt{Clutter-Bench}} establishes a standardized protocol for evaluating methods under increasingly challenging clutter conditions.

In summary, our contributions are threefold:
\begin{itemize}
    \item We propose \textbf{AdaClearGrasp}, a closed-loop framework that models clutter clearing as adaptive high-level planning. It integrates VLM-based semantic reasoning with structured skill invocation and failure feedback, translating high-level decisions into executable atomic actions for robust zero-shot target grasping in clutter.

    \item We introduce \textbf{GeoGrasp}, an object-agnostic RL-based dexterous grasping policy based on relative hand-object representations. This geometry-aware formulation reduces scene overfitting and enables zero-shot grasping across objects with diverse shapes.

    \item We present \textbf{\texttt{Clutter-Bench}}, a standardized simulation benchmark with graded difficulty for evaluating language-conditioned target grasping in clutter. Experiments show that AdaClearGrasp achieves state-of-the-art performance in simulation and zero-shot sim-to-real transfer on real hardware.
\end{itemize}

%% file: section/3_related_works.tex
\section{RELATED WORK}

\subsection{Dexterous Grasping}

Dexterous grasping has evolved from analytical force-closure models to learning-based approaches that generalize across diverse object geometries~\cite{xu2024dexterous, lu2024ugg}. Methods such as UniDexGrasp~\cite{xu2023unidexgrasp} and DexGraspNet~\cite{wang2022dexgraspnet} leverage large-scale datasets and reinforcement learning (RL) to learn robust grasping policies, while D(R,O)-Grasp~\cite{wei2024drograsp} introduces a unified representation of robot–object interactions for cross-embodiment grasp generation. More recent work further improves RL robustness with geometry-aware representations. For example,~\cite{zhang2024graspxl} and RobustDexGrasp~\cite{zhang2025robust} incorporate distance-based features between hand links and graspable object regions, achieving strong sim-to-real transfer via a teacher–student framework.
Despite these advances, dexterous grasping in densely cluttered environments remains difficult due to complex object interactions and occlusions. Existing approaches largely treat manipulation as a low-level control problem and rely on emergent behaviors to handle clutter, without explicitly reasoning about scene structure. In contrast, AdaClearGrasp combines geometry-aware RL with high-level reasoning to determine when and how surrounding objects should be cleared before grasping, enabling more robust dexterous manipulation in cluttered environments.

\subsection{Manipulation in Cluttered Environments}

Manipulating objects in dense clutter is challenging due to physical interference, visual occlusions, and unstable contacts. Early methods typically rely on open-loop grasp pose generation and planning~\cite{wei2022dvgg, zhang2024dexgraspnet, fanganydexgrasp}, offering limited adaptability in dynamic clutter. More recent approaches extend dexterous grasping to handle complex hand–object interactions in cluttered scenes, with reinforcement learning methods such as DexSingRasp~\cite{xu2025dexsingrasp} and ClutterDexGrasp~\cite{chen2025clutterdexgrasp} showing promising results. However, these methods mainly focus on low-level control and emergent behaviors to resolve interference, lacking explicit semantic reasoning for object clearance or scene understanding.
In contrast, AdaClearGrasp leverages vision-language models (VLMs) to introduce higher-level semantic reasoning, enabling the system to decide when and how to clear surrounding objects. By integrating VLM reasoning with atomic dexterous manipulation skills, AdaClearGrasp provides a more adaptive and efficient approach to manipulation in cluttered environments.

\begin{figure*}[t!]
    \centering
    \includegraphics[width=\linewidth]{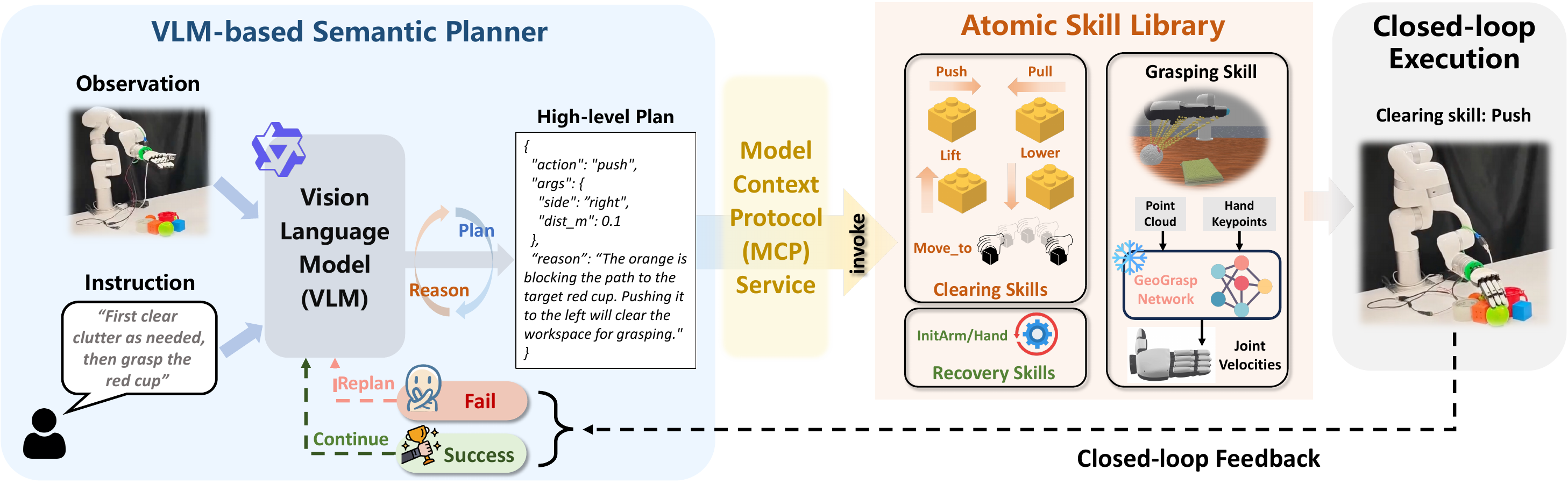}
    \caption{\textbf{Overview of AdaClearGrasp.} 
    The system adopts a hierarchical architecture in which a VLM-based Semantic Planner (left) reasons about the scene and generates high-level action plans. These plans are translated into atomic skills via the Model Context Protocol (MCP) Service (middle). The Skill Library contains heuristic clearing primitives, recovery mechanisms, and a grasping skill (the pretrained geometry-aware RL policy GeoGrasp). The system operates in a closed loop (right), continuously monitoring execution feedback to trigger replanning or recovery when necessary.}
    \label{fig:overview}
    \vspace{-5mm}
\end{figure*}

\subsection{Vision-Language Models for Robotic Manipulation}

Recent work has explored the use of vision-language models (VLMs) for robotic manipulation and planning. Early research demonstrates that VLMs can translate high-level instructions into executable actions or spatial reasoning representations~\cite{ahn2022saycan,huang2023voxposer}. Building on these foundations, recent frameworks have extended VLM reasoning to more complex tasks, such as dexterous manipulation, by integrating perception, language, and control through vision-language-action policies~\cite{liu2025robodexvlm,de2025scaffolding,zhong2025dexgraspvla}. 
While these methods have advanced the field, many still rely on open-loop execution and struggle in cluttered environments with occlusions and unexpected interactions. AdaClearGrasp addresses these limitations by coupling VLM reasoning with atomic skills via a Model Context Protocol (MCP)~\cite{hou2025model}, enabling closed-loop and robust execution with visual feedback and recovery-driven replanning in cluttered environments.

%% file: section/4_method.tex
\section{Method}
\label{sec:method}

We propose \textbf{AdaClearGrasp}, a hierarchical framework for robust dexterous grasping in densely cluttered environments. As illustrated in Fig.~\ref{fig:overview}, the system operates in a closed loop: (1) a Vision-Language Model (VLM) planner analyzes the scene and generates a high-level action plan; (2) the plan is translated into parameterized atomic skills through the Model Context Protocol (MCP); (3) a skill library consisting of heuristic clearing primitives, recovery primitives, and the RL-based \textbf{GeoGrasp} policy executes the interactions; and (4) visual feedback triggers replanning when failures occur.

\subsection{Problem Formulation}

We formulate target grasping in clutter as a Hierarchical Partially Observable Markov Decision Process (POMDP), defined by $(S, A, T, R, O, \gamma)$, with decision making at two levels. At step $t$, the VLM planner receives an observation $o_t \in O$ consisting of the current image $I_t$, the language goal $L$, and execution feedback $f_{t-1}$. It outputs a high-level semantic action $a_{high} \in A_{high}$ to simplify the scene or accomplish the task, where the high-level action space is defined as $A_{high} = A_{clear} \cup A_{grasp}$. Here, $A_{clear} = \{\textit{push}(s,d), \textit{pull}(s,d), \textit{move\_to}(obj), \dots\}$ represents parameterized clearing primitives, and $A_{grasp} = \{\textit{grasp}()\}$ denotes the action to grasp the target object.

Once a high-level action $a_{high}$ is selected, a corresponding low-level policy $\pi_{low}(a_{low}\mid s_t, a_{high})$ is executed. For clearing actions ($a_{high} \in A_{clear}$), $\pi_{low}$ acts as a geometric motion planner that generates deterministic trajectories. For grasping actions ($a_{high} \in A_{grasp}$), $\pi_{low}$ invokes the learned RL policy \textbf{GeoGrasp}, which outputs continuous control commands $a_{low} \in \mathbb{R}^{N_{dof}}$ to execute a stable grasp.

The overall objective is to generate a sequence of high-level actions $a_{high}^{(1)}, a_{high}^{(2)}, \dots, a_{high}^{(T)}$ that grasp and lift the target object within a horizon $T_{max}$. Achieving this requires the planner to reason about occlusions and select appropriate clearing actions before the final grasp.

\begin{figure*}[t!]
    \centering
    \includegraphics[width=0.85\linewidth]{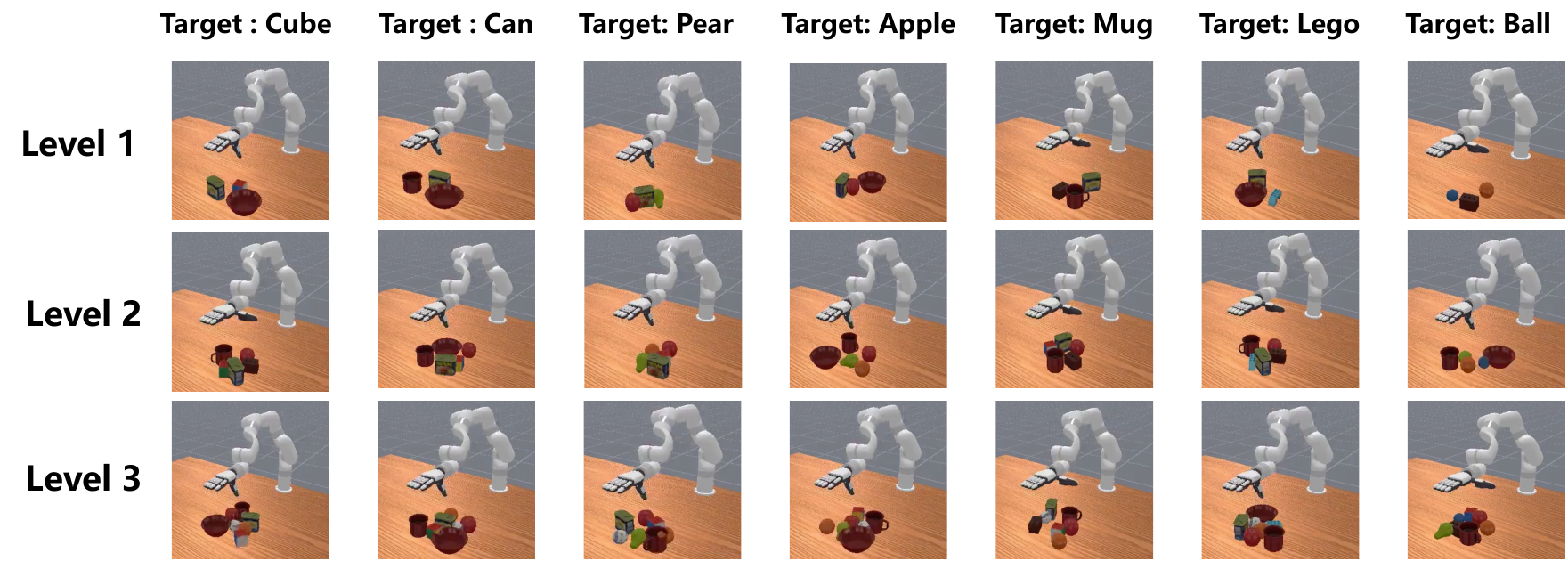}
    \caption{A visualization example of each target object in \textbf{\texttt{Clutter-Bench}} and its corresponding three difficulty levels.}
    \label{fig:clutter_bench}
    \vspace{-5mm}
\end{figure*}

\subsection{VLM-based Semantic Planning}

The high-level reasoning module utilizes a pretrained VLM, Qwen3-VL-32B-Instruct~\cite{bai2025qwen3}, to interpret visual observations, user instructions, and execution feedback. To ensure reliable decision making, we impose structured safety constraints on the planner, including a closed observation space, bounded action parameters (i.e., a restricted action space), and verifiable execution feedback.

\begin{itemize}
    \item \textbf{Observation Space:} The VLM receives an RGB image of the current scene together with a textual list of detected objects, providing both visual and symbolic context for reasoning.
    
    \item \textbf{Action Space:} The model outputs a structured JSON object containing the \textit{action} name (e.g., \textit{push}, \textit{pull}, \textit{grasp}), \textit{args} (parameters such as \textit{side="left"}), and a \textit{reason} field for interpretability.
    
    \item \textbf{Feedback Mechanism:} The prompt includes execution feedback from the previous step (e.g., ``stuck detected'' or ``target not reached''), allowing the VLM to adapt its strategy dynamically, such as switching from \textit{push} to \textit{pull} if the previous action fails.
\end{itemize}

\subsection{Model Context Protocol (MCP) Server}

To bridge the semantic gap between high-level VLM reasoning and low-level robot control, we structure AdaClearGrasp around the Model Context Protocol (MCP), which exposes robot capabilities as callable ``tools'' for the VLM. This allows the planner to invoke manipulation skills through a unified interface.
Under this design, manipulation primitives (e.g., grasp, push, clear) are encapsulated as model-agnostic tool interfaces rather than model-specific function calls, decoupling high-level decision making from hardware-specific implementations and enabling modular skill expansion.
At runtime, when the VLM generates a tool invocation (e.g., \textit{tool\_use: push(side="left")}), the MCP server parses the request and dispatches it to the corresponding skill executor in the \textit{exec/skills} module for motion planning and control execution.
By organizing skills through MCP, AdaClearGrasp supports replacing the VLM, extending the skill library, or migrating to new hardware platforms without modifying the high-level reasoning logic, moving the system closer to a reusable manipulation framework rather than a tightly coupled single-model pipeline.

\subsection{Atomic Skill Library}
Our framework provides a set of parameterized primitives for obstacle clearing, system recovery, and robust grasping.

\subsubsection{Clearing and Recovery Primitives}
We implement deterministic skills based on geometric motion planning to manipulate obstacles and recover from failures.

\begin{itemize}
    \item \textbf{Push \& Pull:} These primitives create space by approaching an object from a specified \textit{side} (\textit{left}, \textit{center}, \textit{right}) relative to its centroid. The end-effector orientation is aligned with the motion direction (e.g., $+\pi/6$ yaw offset for \textit{left}) to improve stability.
    
    \item \textbf{Move \& Lift/Lower:} Cartesian motions such as \textit{move\_to(xy)} and \textit{lift(z)} enable precise repositioning of the end-effector during clearing operations.
    
    \item \textbf{Recovery (InitArm/InitHand):} These primitives reset the arm and hand to safe home configurations when singularities or persistent collisions are detected.
\end{itemize}

\subsubsection{GeoGrasp: RL-based Dexterous Grasping Policy} 
For target acquisition, we introduce \textbf{GeoGrasp}, a dexterous RL policy built on a 59-dimensional geometry-aware observation space that captures local hand–object relations. Rather than pursuing increasingly complex object representations, GeoGrasp focuses on category-agnostic geometric interactions between the hand and object. By grounding the policy in local directional geometry instead of appearance cues, the representation becomes largely insensitive to texture, material properties, and moderate dynamics discrepancies, reducing the sim-to-real gap. This design encourages the policy to reason about geometric reachability and contact structure, leading to stronger zero-shot generalization and more stable real-world transfer.

\paragraph{Observation Space} 
The key design is a \textbf{geometry-aware observation space} ($S \in \mathbb{R}^{59}$) that encodes local hand-object relationships without object category identifiers. It consists of: (1) \textbf{Hand-Object Relations ($O_{geom} \in \mathbb{R}^{54}$)}: unit-normalized nearest-neighbor vectors from 18 hand keypoints to the object point cloud; (2) \textbf{Target State ($O_{target} \in \mathbb{R}^{1}$)}: the target height (Z-coordinate); and (3) \textbf{End-Effector State ($O_{tcp} \in \mathbb{R}^{4}$)}: TCP height and orientation (RPY), providing proprioceptive feedback on the hand pose. 

\paragraph{Reward Function} 
We design a dense reward to guide the policy through reaching, grasping, and lifting. The total reward $R_t$ at time step $t$ is defined as
\begin{equation} 
    R_t = R_{lift} + R_{success} + R_{contact} + R_{nn} - C_{action}, 
\end{equation} 

where $R_{lift} = w_{lift} (h_{obj}^{(t)} - h_{obj}^{(0)})$ encourages vertical displacement from the initial object height, with $h_{obj}^{(t)}$ and $h_{obj}^{(0)}$ denoting the Z-coordinate of the object's center of mass at time $t$ and step $0$.
$R_{success} = w_{success} \mathbb{I}(h_{obj}^{(t)} > 0.15\text{m})$ provides a sparse lifting bonus.
$R_{contact} = w_{contact} N_{contact}$ encourages stable grasping, where $N_{contact}$ is the number of finger links in contact with the object (contact force $>0.5$N).
$R_{nn} = w_{nn} (D_{nn}^{(t-1)} - D_{nn}^{(t)})$ is a shaping reward based on the decrease in nearest-neighbor distance, where
$D_{nn}^{(t)} = \sum_{k=1}^{K} \min_{p \in P_{obj}} \|x_k - p\|_2$
denotes the total distance from $K=18$ hand keypoints $\{x_k\}$ to the object point cloud $P_{obj}$.
$C_{action} = w_{action} \|a_t\|_2$ penalizes large actions to discourage jerky motions.
The weights are $w_{lift}=50.0$, $w_{success}=200.0$, $w_{contact}=10.0$, $w_{nn}=10.0$, and $w_{action}=0.03$. Rewards are clipped to $[-100, 100]$ for training stability.

\paragraph{Training and Deployment} 
The policy (MLP) is trained using PPO~\cite{schulman2017proximal} with 400 parallel environments ($\gamma=0.96$, batch size = 800). During deployment, an IK solver sets the hand to a pre-grasp pose, after which GeoGrasp executes the final grasp. 
The policy is trained on three objects (Cube, Cup, and Apple) in a clutter-free environment. Despite this limited set, GeoGrasp generalizes effectively to unseen geometries. In particular, it successfully grasps the remaining four unseen object categories in \textbf{\texttt{Clutter-Bench}} zero-shot and transfers directly to real robotic hardware without fine-tuning, demonstrating the robustness of the geometry-aware formulation.

\subsection{Closed-loop Execution}
In each operational cycle, the VLM monitors the state through visual feedback. If the target is obstructed, the planner selects a clearing skill (e.g., \textit{pull} the occluding object) to remove the blockage. Once the path is clear, it invokes the \textit{grasp} skill (pre-trained GeoGrasp model) to retrieve the target.
The system also includes a failure recovery mechanism. If a skill fails (e.g., gripper slip, being stuck, or pathfinding failure), execution feedback is returned to the VLM. The planner then generates a new strategy, such as repositioning the arm with \textit{move\_to}, trying another clearing direction, or resetting the manipulator with \textit{initarm}. This cycle continues until the target is grasped and lifted or the maximum step limit is reached.

\subsection{Clutter-Bench}
\label{sec:benchmark}

To systematically evaluate goal-conditioned dexterous grasping in densely cluttered environments, we introduce \textbf{\texttt{Clutter-Bench}}, a benchmark built on ManiSkill3. Existing benchmarks often lack standardized clutter configurations or isolate grasping from high-level reasoning and scene interaction, making it difficult to evaluate integrated capabilities such as reasoning, clearing, and manipulation. \textbf{\texttt{Clutter-Bench}} addresses this gap through a graded and fully reproducible evaluation protocol.

As illustrated in Fig.~\ref{fig:clutter_bench}, \textbf{\texttt{Clutter-Bench}} covers diverse object geometries and clutter densities:
\textbf{(1) Object Diversity:} The benchmark includes seven target objects from the YCB dataset, covering a broad range of shapes and physical properties. Ordered by increasing grasping difficulty in our experiments, they are: \textbf{cube} (box-shaped), \textbf{can} (cylindrical), \textbf{pear} (irregular), \textbf{apple} (spherical), \textbf{mug} (hollow with handle), \textbf{lego} (complex geometry), and \textbf{ball} (small spherical, challenging due to rolling dynamics).
\textbf{(2) Graded Difficulty Tiers:} Three difficulty levels are defined by the number of obstacles: \textbf{Level-1} (2 obstacles), \textbf{Level-2} (4 obstacles), and \textbf{Level-3} (6 obstacles). For each object–difficulty pair, we pre-generate \textbf{10 scene configurations}. Object positions are sampled within a $20 \times 20$ cm$^2$ region around the target, with orientations uniformly drawn from $[0^{\circ}, 360^{\circ})$. 
All configurations are stored as JSON files specifying target and obstacle poses ($x, y, \theta$), ensuring identical evaluation layouts across methods and eliminating randomness in scene initialization. 
Obstacles are sampled from common household items and placed nearby to induce occlusion.

%% file: section/5_exp.tex
\begin{figure}[t]
    \centering
    
    \includegraphics[width=0.9\linewidth]{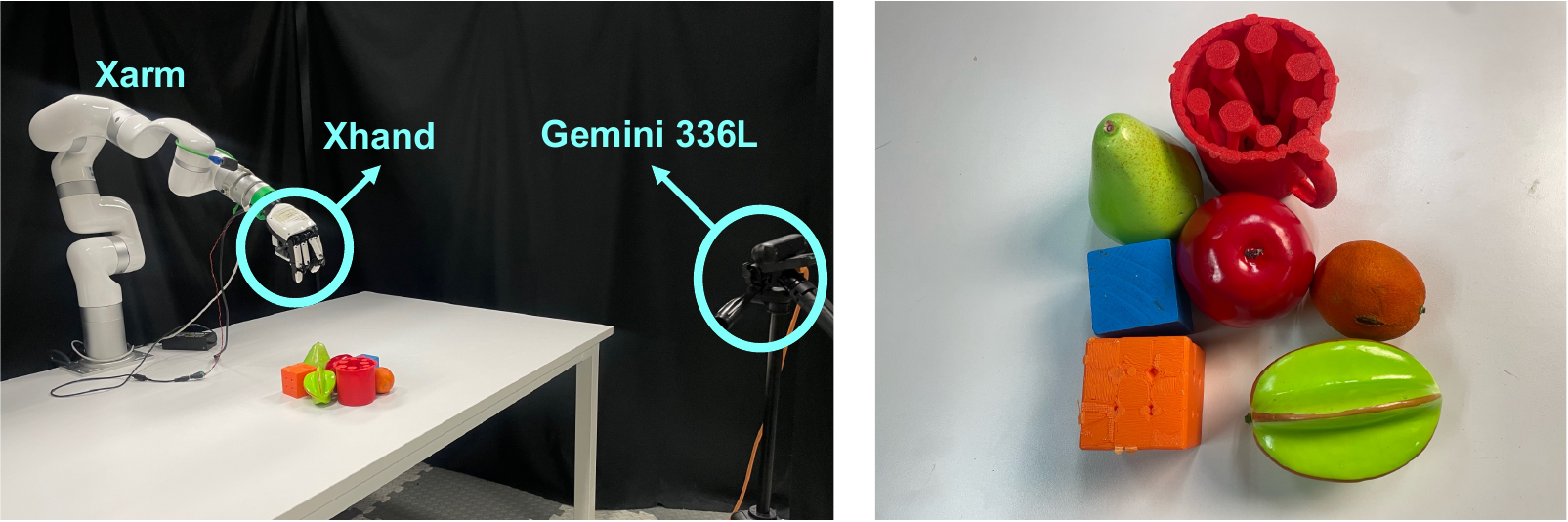}
    
    \vspace{1mm}
    
    \includegraphics[width=0.9\linewidth]{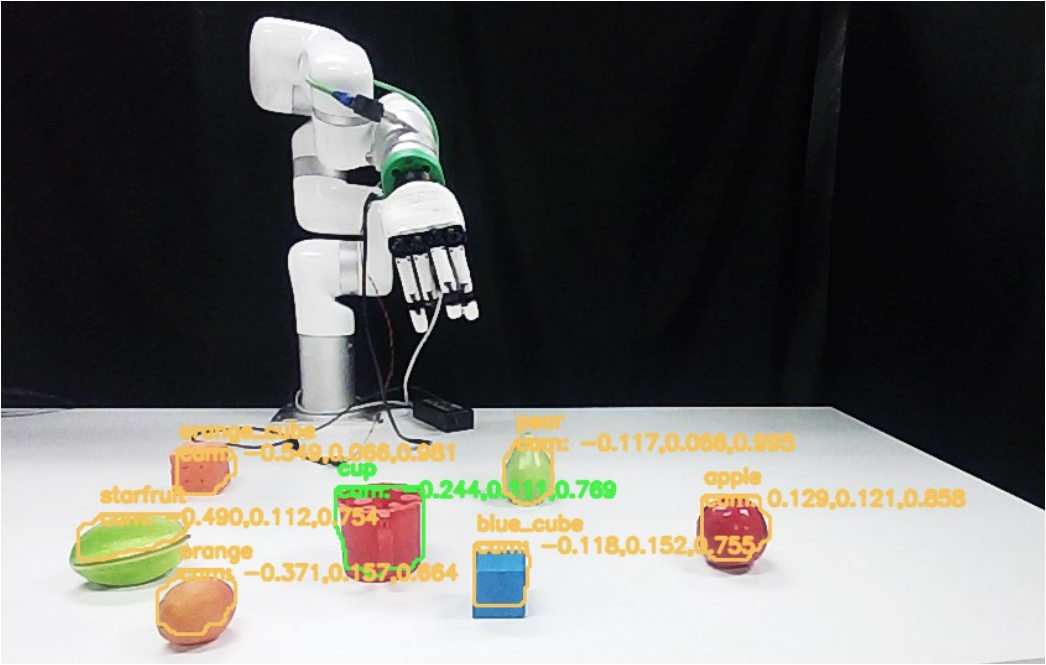}
    
    \caption{\textbf{Real-world Implementation Details.} 
    (Upper) Real-world experimental setup with xArm7, XHand, and the external camera Gemini 336L. 
    (Lower) Visualization of 6D pose estimation using FoundationPose, which provides input to the VLM planner.}

    \label{fig:real_world_combined}
    \vspace{-3mm}
\end{figure}

\begin{figure}
    \centering
    \includegraphics[width=0.95\linewidth]{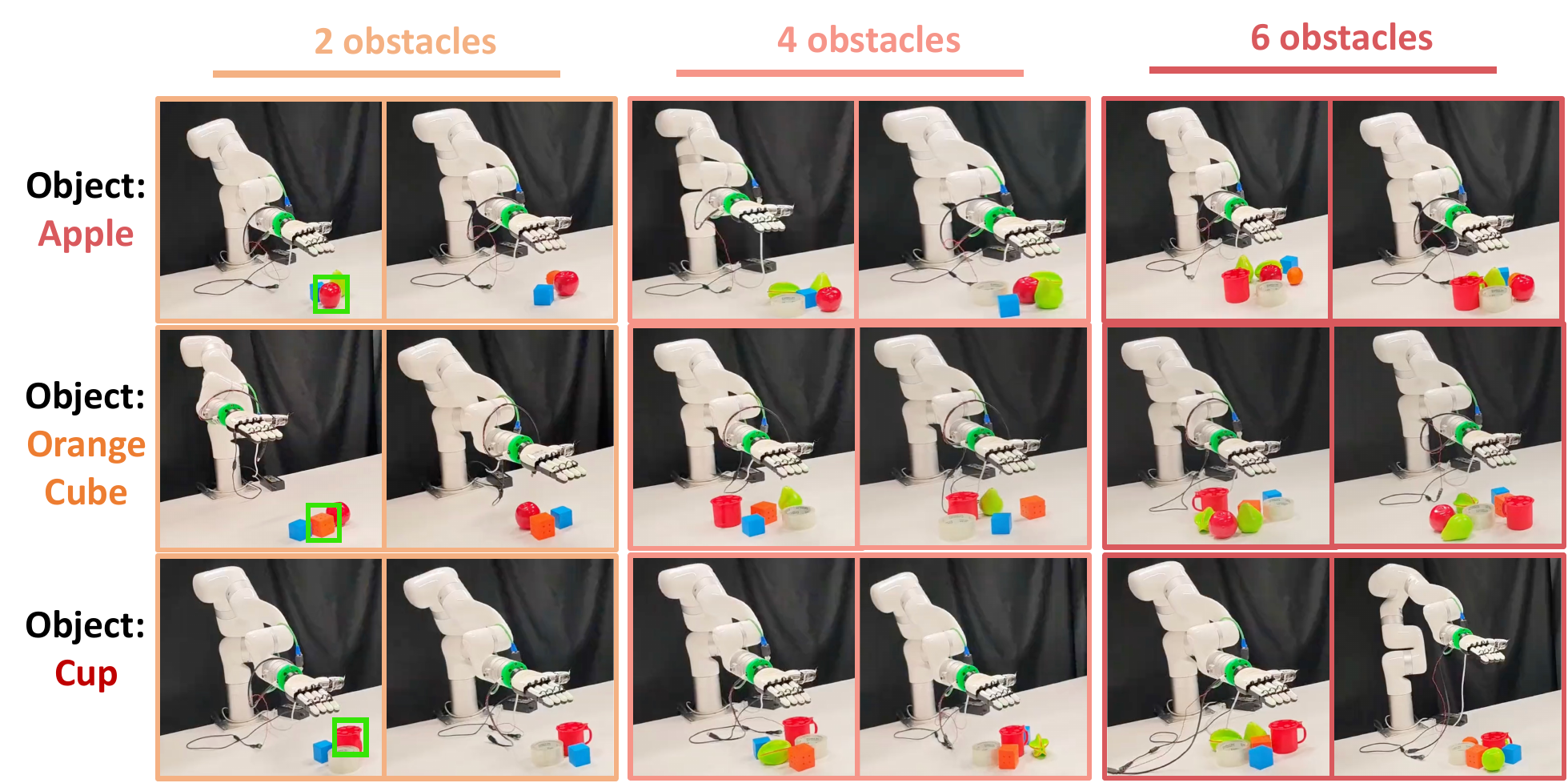}
    \caption{Real-world experiments on 18 diverse scenarios.}
    \label{fig:real_task}
    \vspace{-5mm}
\end{figure}

\section{EXPERIMENTS}
\label{sec:experiments}
In this section, we evaluate AdaClearGrasp in both simulation and the real world to answer four key research questions:

{\setlength{\leftmargini}{8pt}
\begin{itemize}
    \item \textbf{RQ1:} Can AdaClearGrasp achieve robust zero-shot dexterous grasping across varying difficulty levels in cluttered simulation environments?
    \item \textbf{RQ2:} How do the adaptive clearing strategy and closed-loop feedback mechanism contribute to task success?
    \item \textbf{RQ3:} How effectively does the GeoGrasp policy, trained on a single object, generalize to diverse unseen object geometries in a zero-shot manner? 
    \item \textbf{RQ4:} Can AdaClearGrasp perform Sim-to-Real transfer for dense clutter grasping tasks in the physical world without fine-tuning?
\end{itemize}
}

\input{tables/main_table}

\subsection{Experimental Setup}
\label{sec:setup}

We conduct experiments in both a high-fidelity physics simulator and on a real-world robotic platform.

\paragraph{Simulation Setup} 
Our simulation environment is built on ManiSkill3, with an xArm7 robot equipped with a dexterous XHand. A front-mounted RGB-D camera provides $128 \times 128$ observations. In \textbf{\texttt{Clutter-Bench}}, we randomly place 2, 4, or 6 obstacles from 7 YCB categories (Sec.~\ref{sec:benchmark}) within a $20\text{cm} \times 20\text{cm}$ region around the target.

We compare AdaClearGrasp with three baselines:
(1) \textit{VLM Scaffolding}~\cite{de2025scaffolding}: A framework that uses VLM-predicted keypoints and 3D trajectories to guide fine-grained manipulation. Designed primarily for single-object interaction, it lacks explicit obstacle-handling mechanisms, highlighting the robustness of our approach in dense clutter.
(2) \textit{GeoGrasp (No VLM):} Directly executes the GeoGrasp policy without any clearing actions, evaluating the importance of VLM-guided clearing for creating feasible grasp paths.
(3) \textit{AdaClearGrasp (w/o Replan):} An ablated version that follows the initial plan without replanning upon failure, isolating the contribution of the closed-loop feedback mechanism.

For implementation, the VLM planner uses Qwen3-VL-32B-Instruct~\cite{bai2025qwen3}. The GeoGrasp policy (a 3-layer MLP with hidden sizes $[256,256,256]$) is trained with PPO for 6 million steps using 400 parallel environments ($\gamma=0.96$, batch size=800). All experiments are conducted on an NVIDIA RTX 5080 GPU and an Intel Core i9-14900K CPU.

\paragraph{Real-world Setup}
To evaluate the Sim-to-Real transfer capability of AdaClearGrasp, we replicate the simulation setup on a physical platform shown in Fig.~\ref{fig:real_world_combined}. The system consists of an xArm7 manipulator equipped with an XHand. For perception, we employ FoundationPose~\cite{wen2024foundationpose} to estimate the 6D poses of all objects in the scene. The estimated poses, together with the RGB image, are provided to Qwen3-VL-32B-Instruct for high-level semantic planning. The low-level grasping skill uses the GeoGrasp policy transferred directly from simulation without fine-tuning, leveraging its geometry-aware formulation to mitigate the reality gap.
To ensure a rigorous evaluation, we construct a real-world benchmark that mirrors the \textbf{\texttt{Clutter-Bench}} protocol. As shown in Fig.~\ref{fig:real_task}, it includes 18 scenarios covering three representative target objects (Apple, Orange Cube, and Cup) across three clutter levels (2, 4, and 6 obstacles). These scenarios incorporate diverse obstacles to challenge system robustness in real-world environments.

\paragraph{Metric}
We evaluate performance using the \textbf{Success Rate (SR)}, defined as the percentage of trials in which the target object is successfully grasped and lifted to a height of at least 15 cm, without dropping, for 2 seconds. A trial is considered a failure if the robot fails to grasp the target within 40 planning steps or if the target or obstacles are knocked out of the workspace.

\subsection{Zero-shot Performance on \textbf{\texttt{Clutter-Bench}} (RQ1)}
We evaluate the zero-shot dexterous grasping capability of AdaClearGrasp against the VLM Scaffolding baseline across the three difficulty tiers of \textbf{\texttt{Clutter-Bench}}. Table~\ref{tab:comparison_main} reports the quantitative success rates. Additional simulation demonstrations are available in the supplementary video.

\textbf{Robustness in Dense Clutter.} 
The results demonstrate the strong capability of AdaClearGrasp in highly cluttered environments. While the baseline achieves only a 6\% success rate in the sparse Level-2 setting, its performance drops to 0\% at Level-4 and Level-6 as clutter density increases. This failure arises because the baseline models manipulation as a sequence of collision-free waypoints, which rarely exist in dense scenes due to severe occlusions. In contrast, AdaClearGrasp maintains success rates of \textbf{89\%}, \textbf{84\%}, and \textbf{76\%} across the three levels. These results highlight the importance of the adaptive clearing strategy for enabling grasping when the workspace is heavily constrained by surrounding objects.

\textbf{Handling Diverse Physical Properties.} AdaClearGrasp demonstrates consistent performance across objects with diverse physical characteristics. For stable geometries like \textit{cubes} and \textit{cans}, the method achieves near-perfect success rates ($\ge 90\%$). Even for objects with challenging dynamics—such as rolling \textit{balls} or irregular \textit{legos}—the system maintains success rates of 40–70\%. This suggests that the adaptive clearing strategy effectively isolates the target without excessive disturbance, creating a safe workspace for subsequent precision grasp.

\begin{figure}
    \centering
    \includegraphics[width=\linewidth]{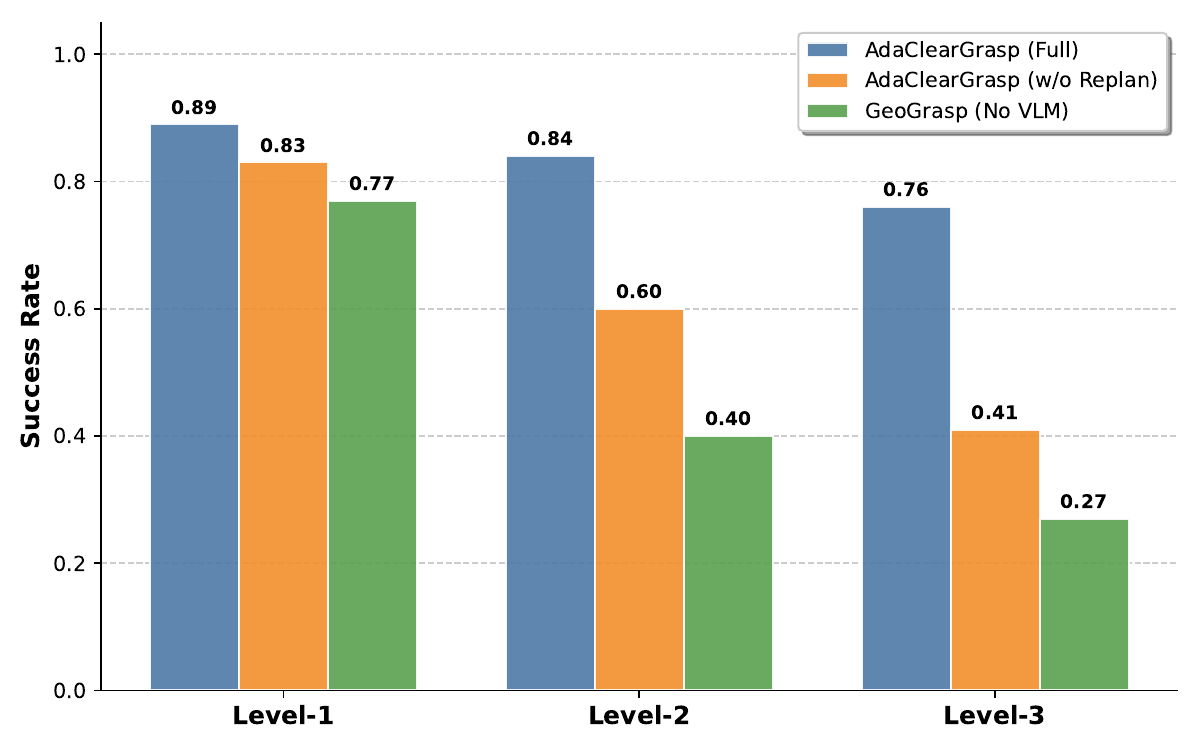}
    \caption{Ablation study on adaptive clearing and closed-loop feedback.}
    \label{fig:ablation}
    \vspace{-6mm}
\end{figure}

\subsection{Ablation Study (RQ2)}
To analyze the contributions of the adaptive clearing strategy and the closed-loop feedback mechanism in AdaClearGrasp, we evaluate ablated variants across all difficulty levels. Fig.~\ref{fig:ablation} summarizes the results.

\textbf{Effectiveness of Adaptive Clearing.} 
Comparing the full method with the \textit{GeoGrasp (No VLM)} baseline highlights the importance of obstacle clearing. In the sparse Level-1 setting, direct grasping achieves a 77\% success rate, but performance drops to 40\% (Level-2) and 27\% (Level-3) as clutter increases. This suggests that in dense scenes the target is often physically inaccessible without rearranging surrounding objects. In contrast, the VLM-guided adaptive clearing strategy creates feasible grasp affordances, maintaining high success rates even under heavy clutter.

\textbf{Importance of Closed-loop Feedback.} 
The comparison between the full method and the \textit{w/o Replan} variant demonstrates the importance of closed-loop execution. Although the initial clearing plan performs reasonably well in Level-1 (83\%), the absence of replanning leads to substantial degradation in more complex scenes, dropping to 41\% at Level-3. By enabling dynamic replanning for up to five attempts after execution failures, the full system significantly outperforms the open-loop variant (e.g., +35\% at Level-3). This mechanism improves robustness to execution errors and environmental uncertainty, allowing recovery from failed grasps or incomplete clearing actions.

\begin{figure}
    \centering
    \includegraphics[width=0.8\linewidth]{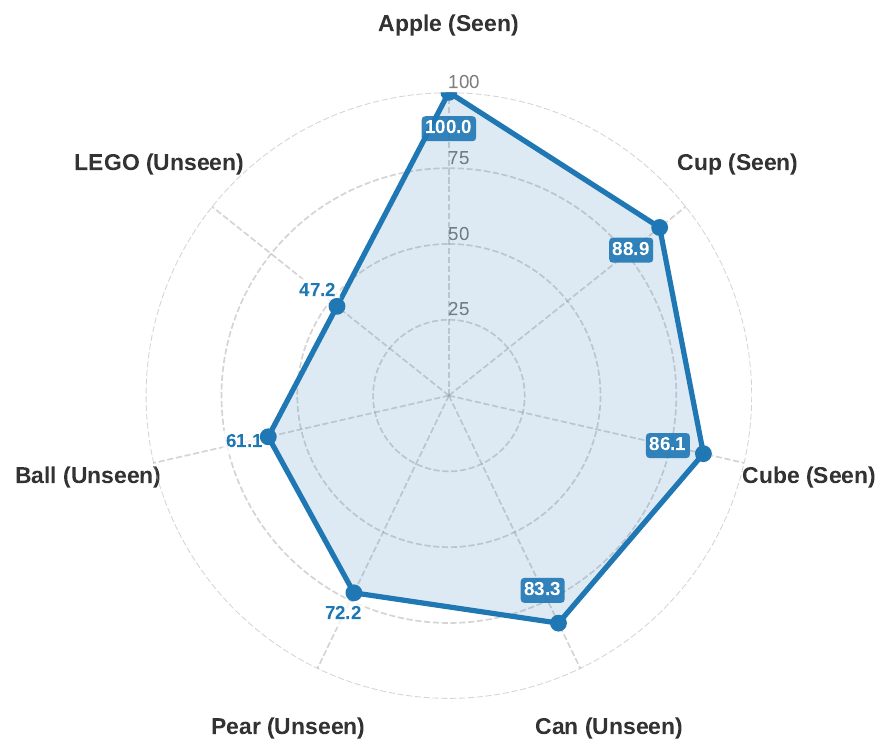}
    \caption{The grasp success rates of our GeoGrasp policy on 7 object categories (3 seen, 4 unseen) in clutter-free environments.}
    \label{fig:grasp_success_rates}
    \vspace{-5mm}
\end{figure}

\subsection{Generalization Analysis of GeoGrasp (RQ3)}
To evaluate the intrinsic generalization capability of our GeoGrasp policy, we conduct isolated grasping experiments in a clutter-free environment. As shown in Fig.~\ref{fig:grasp_success_rates}, we assess performance across 7 object categories: 3 seen during training and 4 novel, unseen objects. To rigorously test grasping robustness, we randomly sample initial object positions within a $20 \times 20$ disturbance area and vary orientations uniformly over $[0^{\circ}, 360^{\circ})$. Qualitative visualizations of the GeoGrasp policy can be found in the supplementary video.

\textbf{Performance on Seen Objects.} 
The policy excels on the training set, achieving high success rates for the \textit{Cube} (86.1\%), \textit{Cup} (88.9\%), and perfect performance on the \textit{Apple} (100.0\%). This confirms that GeoGrasp learns robust grasping strategies for objects with varying shapes (box, cylinder, sphere) and sizes.

\textbf{Generalization to Unseen Geometries.} 
The policy shows strong generalization to novel geometries. It performs well on the \textit{Can} (83.3\%) and \textit{Pear} (72.2\%), suggesting that learned geometric features (e.g., local curvature, antipodal points) generalize across semantically different but geometrically related objects. Even for challenging cases like the rolling \textit{Ball} (61.1\%) and non-convex \textit{LEGO} (47.2\%), GeoGrasp achieves reasonable success without fine-tuning. This validates the effectiveness of our geometry-centric observation space in capturing universal geometric cues necessary for stable grasping beyond the training distribution.

\input{tables/real_grasp}

\subsection{Real-world Sim-to-Real Transfer (RQ4)}
To answer RQ4, we systematically evaluate AdaClearGrasp in the physical world across 90 trials. As shown in Fig.~\ref{fig:real_task}, our test matrix covers 3 target objects, each tested at 3 clutter levels with 2 distinct scene configurations per level, repeated for 5 trials per scene. Table~\ref{tab:real_world} summarizes the results. More real-world demos are available on the anonymous project website and in the supplementary video.

\textbf{Successful Zero-shot Transfer.} 
Despite the significant reality gap—including sensor noise from FoundationPose, unmodeled friction, and physical contact dynamics—our system achieves a respectable overall success rate of \textbf{70.0\%} (63/90). This confirms the effectiveness of our simulation-based training pipeline and the robustness of the GeoGrasp policy, which transfers directly to the real robot without fine-tuning. Among the targets, the \textit{Orange Cube} performed best (80.0\% SR), likely due to its stable flat surfaces.

\textbf{Impact of Clutter Density.} 
Performance shows a clear trend across clutter levels. In the sparse Level-2 setting, AdaClearGrasp achieves a high success rate of \textbf{90.0\%}. As the environment becomes more cluttered, performance drops to \textbf{70.0\%} (Level-4) and \textbf{50.0\%} (Level-6). This degradation is primarily due to the increased complexity of the physical configuration space. At Level-6, narrow margins for error, combined with pose estimation jitters or slight slippage during clearing, lead to collisions or failed grasps.

\textbf{Failure Analysis.} 
Failures in dense scenes were mainly due to real-world uncertainties. For example, in several Level-6 trials, the \textit{Cup} was knocked over while clearing adjacent obstacles due to unpredictable contact forces. Nonetheless, the system successfully retrieved the target in half of the most extreme scenarios, demonstrating the viability of AdaClearGrasp for handling real-world dense clutter.

%% file: tables/main_table.tex
\begin{table*}[t!]
\centering
\caption{Success Rates Comparison on \texttt{\textbf{Clutter-Bench}}. We evaluate the zero-shot performance of AdaClearGrasp against the VLM Scaffolding baseline across three increasing difficulty tiers (Level-1 to Level-3). The results are averaged over 10 independent trials for each target object and clutter configuration.}
\label{tab:comparison_main}
\small
\renewcommand{\arraystretch}{0.8}

\begin{tabular}{c | c c | c c | c c}
\toprule
\multirow{2}{*}{\textbf{Target Object}} 
& \multicolumn{2}{c|}{\textbf{Level-1}} 
& \multicolumn{2}{c|}{\textbf{Level-2}} 
& \multicolumn{2}{c}{\textbf{Level-3}} \\
\cmidrule(lr){2-3} \cmidrule(lr){4-5} \cmidrule(lr){6-7}

& VLM Scaffolding 
& \cellcolor{violet!6}\textbf{AdaClearGrasp} 
& VLM Scaffolding 
& \cellcolor{violet!6}\textbf{AdaClearGrasp} 
& VLM Scaffolding 
& \cellcolor{violet!6}\textbf{AdaClearGrasp} \\

\midrule

\cellcolor{blue!3}Cube   & 0.2 & \cellcolor{violet!6}\textbf{0.8} & 0.0 & \cellcolor{violet!6}\textbf{0.9} & 0.0 & \cellcolor{violet!6}\textbf{1.0} \\

\cellcolor{blue!4}Can    & 0.1 & \cellcolor{violet!6}\textbf{0.9} & 0.0 & \cellcolor{violet!6}\textbf{1.0} & 0.0 & \cellcolor{violet!6}\textbf{0.9} \\

\cellcolor{blue!5}Pear   & 0.0 & \cellcolor{violet!6}\textbf{1.0} & 0.0 & \cellcolor{violet!6}\textbf{1.0} & 0.0 & \cellcolor{violet!6}\textbf{0.7} \\

\cellcolor{blue!6}Apple  & 0.1 & \cellcolor{violet!6}\textbf{1.0} & 0.0 & \cellcolor{violet!6}\textbf{0.8} & 0.0 & \cellcolor{violet!6}\textbf{0.9} \\

\cellcolor{blue!7}Mug    & 0.0 & \cellcolor{violet!6}\textbf{0.8} & 0.0 & \cellcolor{violet!6}\textbf{0.8} & 0.0 & \cellcolor{violet!6}\textbf{0.9} \\

\cellcolor{blue!8}Lego   & 0.0 & \cellcolor{violet!6}\textbf{1.0} & 0.0 & \cellcolor{violet!6}\textbf{0.7} & 0.0 & \cellcolor{violet!6}\textbf{0.5} \\

\cellcolor{blue!10}Ball  & 0.0 & \cellcolor{violet!6}\textbf{0.7} & 0.0 & \cellcolor{violet!6}\textbf{0.7} & 0.0 & \cellcolor{violet!6}\textbf{0.4} \\

\midrule

\rowcolor{gray!5}
Average & 0.06 & \textbf{0.89} & 0.00 & \textbf{0.84} & 0.00 & \textbf{0.76} \\

\bottomrule
\end{tabular}
\vspace{-6mm}
\end{table*}

%% file: tables/real_grasp.tex
\begin{table}[t]
    \centering
    \caption{Real-world Sim-to-Real Transfer Results.}
    \label{tab:real_world}
    \small
    \setlength{\tabcolsep}{1.3pt}
    \renewcommand{\arraystretch}{1.1}
    \begin{tabular}{l | c c c | c}
        \toprule
        \multirow{2}{*}{\textbf{Target Object}} & \multicolumn{3}{c|}{\textbf{Clutter Level (Obstacles)}} & \multirow{2}{*}{\textbf{Avg.}} \\
        \cmidrule(lr){2-4}
         & \textbf{2} & \textbf{4} & \textbf{6} & \\
        \midrule
        Apple & 9/10 & 7/10 & 5/10 & \textbf{70\%} \\
        Orange Cube & 10/10 & 8/10 & 6/10 & \textbf{80\%} \\
        Cup & 8/10 & 6/10 & 4/10 & \textbf{60\%} \\
        \midrule
        Total Success & 27/30 & 21/30 & 15/30 & \multirow{2}{*}{\textbf{70\%}} \\
        Success Rate & 90\% & 70\% & 50\% & \\
        \bottomrule
    \end{tabular}
    \vspace{-6mm}
\end{table}

%% file: section/6_conclusion.tex
\section{Conclusion}
\label{sec:conclusion}

We present AdaClearGrasp, a closed-loop hierarchical framework for learning adaptive clearing for zero-shot robust grasping in densely cluttered environments. A Vision–Language Model (VLM) reasons over scene semantics and instruction to select among obstacle clearing, direct grasping, and recovery skills, enabling adaptive behavior while balancing safety and efficiency. At the execution level, we introduce GeoGrasp, a geometry-aware reinforcement learning policy that enables stable dexterous grasping with strong zero-shot generalization across diverse object categories. A closed-loop feedback mechanism improves reliability by monitoring task progress and triggering replanning upon failures, allowing the system to recover and operate robustly in dense clutter.
To evaluate our approach, we introduce \textbf{\texttt{Clutter-Bench}}, the first simulation benchmark with graded clutter complexity. Experiments across 210 simulated scenarios and 18 real-world trials show that AdaClearGrasp consistently outperforms baselines and achieves state-of-the-art success rates in densely cluttered environments.
Future work will focus on enhancing the low-level dexterous grasping policy to support cross-embodiment transfer across diverse robotic hands and morphologies.

%% file: section/7_appendix.tex
\section{Appendix}
\label{sec:appendix}

\subsection{GeoGrasp Implementation Details}
\label{app:geograsp}

While the main text outlines the core principles of GeoGrasp, we provide here the specific implementation details necessary for reproduction, focusing on keypoint definitions, action space scaling, and domain randomization parameters used to bridge the sim-to-real gap.

\subsubsection{Keypoint Definition for Geometric Observation}
The 54-dimensional geometric observation $O_{geom}$ relies on 18 specific keypoints on the XHand. These keypoints are strategically distributed to cover the palm and phalanges, which are critical for both power and precision grasps.
\begin{itemize}
    \item \textbf{Palm (6 points):} Distributed across the palm surface to detect proximity for power grasps.
    \item \textbf{Fingers (12 points):} 3 keypoints per finger (Thumb, Index, Middle, Ring), located at the center of the proximal, medial, and distal links.
\end{itemize}
For each keypoint $x_k$, we compute the nearest vector $v_k = p^* - x_k$ to the object point cloud $P_{obj}$. The point cloud $P_{obj}$ consists of $N=1024$ points sampled uniformly from the mesh surface. In simulation, we use ground-truth sampling; in real-world deployment, this is approximated by transforming the canonical object mesh using the pose estimated by FoundationPose.

\subsubsection{Action Space and Control}
The policy outputs a 19-dimensional continuous action vector $a_t \in [-1, 1]^{19}$, which is mapped to joint position targets for the PD controller. To ensure smooth and safe motion, we apply scaling to the raw network outputs:
\begin{itemize}
    \item \textbf{xArm Control (7 dims):} Controlling the 7-DOF arm joints. The raw output is scaled by a factor of 0.05 rad/step.
    \item \textbf{XHand Control (12 dims):} Controlling the 12-DOF hand joints. The raw output is scaled by 0.05 rad/step.
\end{itemize}
We employ a relative control scheme where the target joint position is updated as $q_{target}^{(t)} = q_{current}^{(t)} + \text{scale} \cdot a_t$.

\subsubsection{Domain Randomization}
To enhance the robustness of the policy for Sim-to-Real transfer, we apply extensive domain randomization during training:
\begin{itemize}
    \item \textbf{Physical Properties:} Friction coefficients are randomized uniformly in $[0.5, 2.0]$, and object mass is randomized by $\pm 20\%$.
    \item \textbf{Observation Noise:} Gaussian noise ($\sigma=0.005$m) is added to the object point cloud positions to simulate perception jitter.
    \item \textbf{Initialization:} The robot's initial joint positions are perturbed by $\pm 0.05$ rad to prevent the policy from overfitting to a fixed starting pose.
\end{itemize}

\subsubsection{PPO Hyperparameters}
We use the Stable Baselines3 implementation of PPO with the hyperparameters listed in Table~\ref{tab:ppo_params}.

\begin{table}[h]
    \centering
    \caption{PPO Hyperparameters for GeoGrasp Training}
    \label{tab:ppo_params}
    \small
    \renewcommand{\arraystretch}{1.1} % 稍微增加行高
    \setlength{\tabcolsep}{12pt} % 增加列间距
    \begin{tabular}{l c}
        \toprule
        \textbf{Parameter} & \textbf{Value} \\
        \midrule
        Optimizer & Adam \\
        Learning Rate & $3 \times 10^{-4}$ \\
        Num. Environments & 400 \\
        Rollout Steps per Env & 5 \\
        Batch Size & 800 \\
        Epochs per Rollout & 5 \\
        Discount Factor ($\gamma$) & 0.96 \\
        GAE Lambda ($\lambda$) & 0.95 \\
        Clip Range & 0.2 \\
        Entropy Coefficient & 0.0 \\
        Value Function Coeff & 0.5 \\
        Max Gradient Norm & 0.5 \\
        \bottomrule
    \end{tabular}
\end{table}

\subsection{VLM Planner Implementation Details}
\label{app:vlm_planner}

We provide the complete prompt structure and tool definitions used to interface the VLM planner with the low-level control system via the Model Context Protocol (MCP).

\subsubsection{VLM Prompt Structure}
The prompt fed to Qwen3-VL consists of three components:
\begin{enumerate}
    \item \textbf{System Prompt:} Defines the agent's role, the available toolset, and the strict JSON output format.
    \item \textbf{Task Description:} Provides the semantic context, including the target object name and a list of detected objects in the scene.
    \item \textbf{Dynamic Context:} Contains the current visual observation (RGB image) and the execution feedback from the previous step.
\end{enumerate}
Fig.~\ref{fig:vlm_prompt} illustrates the structured template. We use a ``Chain-of-Thought'' style instruction (``reason'' field) to encourage the VLM to articulate its planning logic before emitting the action.

\subsubsection{MCP Tool Definitions}
Table~\ref{tab:tool_definitions} details the parameterized skills exposed to the VLM. Each tool maps to a deterministic or learning-based policy in the low-level controller.

\begin{table*}[t!]
    \centering
    \caption{Atomic Skills and their Parameters exposed via MCP}
    \label{tab:tool_definitions}
    \small
    \renewcommand{\arraystretch}{1.2}
    \setlength{\tabcolsep}{8pt}
    % 使用 p 列来自动换行，调整宽度以适应两栏布局
    \begin{tabular}{l p{5cm} p{8cm}}
        \toprule
        \textbf{Tool Name} & \textbf{Parameters} & \textbf{Description} \\
        \midrule
        \texttt{push} & \texttt{side} (left, right, center), \texttt{dist} & Approaches the nearest obstacle relative to the target and pushes it away to clear a path. \\
        \texttt{pull} & \texttt{side} (left, right, center), \texttt{dist} & Hooks the nearest obstacle and pulls it closer to the robot base to remove occlusion. \\
        \texttt{pick} & None & Invokes the \textbf{GeoGrasp} policy to attempt a dexterous grasp on the target object. \\
        \texttt{move\_to} & \texttt{target} (string) & Moves the end-effector to a pre-grasp hover position above the specified object. \\
        \texttt{lift} & \texttt{height} (optional) & Lifts the end-effector vertically by a specified distance to avoid collisions during transport. \\
        \texttt{initarm} & None & Resets the arm to a safe home configuration to recover from singularities or execution failures. \\
        \bottomrule
    \end{tabular}
\end{table*}

\begin{figure*}[t!]
    \centering
    % 使用 tcolorbox 增强 Prompt 的展示效果
    \begin{tcolorbox}[
        colback=gray!5!white,
        colframe=gray!75!black,
        title=\textbf{VLM Prompt Template},
        arc=1mm, % 圆角
        boxrule=0.8pt
    ]
    \small \ttfamily % 使用等宽字体
    \textbf{System Prompt:} \\
    You are a careful and disciplined robot manipulation planner. Your goal is to clear obstacles and grasp the target object. \\
    \textbf{Available Tools:} \\
    - \texttt{move\_to(name)}, \texttt{push(side, dist\_m)}, \texttt{pull(side, dist\_m)} \\
    - \texttt{lift() / lower()}, \texttt{grasp()}, \texttt{initarm() / inithand()} \\
    \textbf{Output Format:} Return a strictly valid JSON object: \\
    \texttt{\{ "action": "...", "args": \{...\}, "reason": "..." \}}
    
    \rule{\textwidth}{0.4pt}
    
    \textbf{User Task (Initial):} \\
    \textbf{Objective:} Clear clutter around the target object to enable a successful grasp. \\
    \textbf{Scene Context:} The target object is a \texttt{<TARGET\_NAME>}. \\
    \textbf{Available Objects:} \texttt{[target, obstacle\_1, obstacle\_2, ...]}
    
    \rule{\textwidth}{0.4pt}
    
    \textbf{User Input (Step $t$):} \\
    \textbf{Visual Observation:} \texttt{<IMAGE\_BASE64>} \\
    \textbf{Execution Feedback (from Step $t-1$):} \\
    \texttt{\{ "action": "push", "success": false, "message": "collision detected" \}} \\
    \textbf{Instruction:} Analyze the image and feedback. If the previous action failed, propose an alternative strategy. Output the next action in JSON.
    \end{tcolorbox}
    \vspace{-2mm}
    \caption{The structured prompt template used for VLM planning in AdaClearGrasp. It integrates system instructions, task context, and dynamic feedback to guide the VLM's reasoning.}
    \label{fig:vlm_prompt}
\end{figure*}

\subsubsection{Execution Feedback Handling}
The system feeds back structured execution results to the VLM to close the loop. If an action fails (e.g., ``collision\_detected'' or ``gripper\_slip''), the VLM receives a JSON report containing the error type and a descriptive message, allowing it to adapt its strategy in the subsequent planning step.

\subsection{Clutter-Bench Details}
\label{app:clutter_bench}

To ensure fair and reproducible comparisons, \textbf{\texttt{Clutter-Bench}} relies on a rigorous procedural generation pipeline that creates stable, physically valid, and densely cluttered scenes. We detail the generation protocol below.

\subsubsection{Scene Generation Protocol}
For each target object, we define a \textit{base scene} configuration containing the target model and a pool of 6 candidate clutter objects selected from the YCB dataset. The generation process for a specific difficulty level $N \in \{2, 4, 6\}$ follows these steps:

\begin{enumerate}
    \item \textbf{Object Selection:} The target object and the first $N$ obstacles from the candidate pool are instantiated.
    \item \textbf{Pose Sampling:} We randomly sample the 2D position $(x, y)$ and yaw angle $\psi$ for every object within a confined workspace:
    \begin{itemize}
        \item $x, y \sim \mathcal{U}(-0.10\text{m}, 0.10\text{m})$: The workspace is restricted to a $20 \times 20$ cm$^2$ area to force high-density packing.
        \item $\psi \sim \mathcal{U}(-\pi, \pi)$: Full rotation randomization.
        \item \textbf{Constraint:} A minimum initial distance of 6cm is enforced between object centers to prevent deep interpenetration.
    \end{itemize}
    \item \textbf{Physics Stabilization:} After initialization, objects are dropped from a small height ($z \approx 3$mm) and the simulation is stepped forward to allow physics settling.
    \begin{itemize}
        \item \textbf{Settle Phase:} We run the physics engine for 30 steps ($0.5$s) to let objects fall and resolve contacts.
        \item \textbf{Measurement Phase:} We continue simulation for another 60 steps ($1.0$s) to monitor stability.
        \item \textbf{Rejection Criterion:} If any object's displacement exceeds 1cm during the measurement phase (indicating rolling, sliding, or instability), the scene is discarded and re-sampled.
    \end{itemize}
    \item \textbf{Serialization:} Once a stable configuration is found, the precise poses of all objects are serialized into a JSON file. This guarantees that all baselines are evaluated on \textit{identical} scene states.
\end{enumerate}

\subsubsection{Object Roster}
Table~\ref{tab:clutter_objects} lists the specific YCB model IDs used for targets and their corresponding clutter obstacles. We carefully select clutter objects to ensure a mix of geometries (boxes, spheres, cylinders) and physical properties (friction, mass).

\begin{table}[h]
    \centering
    \caption{Target Objects and Clutter Pools in \textbf{\texttt{Clutter-Bench}}.}
    \label{tab:clutter_objects}
    \small
    \renewcommand{\arraystretch}{1.1}
    \setlength{\tabcolsep}{6pt}
    \begin{tabular}{l l l}
        \toprule
        \textbf{Target Name} & \textbf{YCB Model ID} & \textbf{Typical Clutter Objects} \\
        \midrule
        \textbf{Apple} & \texttt{013\_apple} & Orange, Foam Brick, Mug, Pear, Ball, Cube \\
        \textbf{Cube} & \texttt{077\_rubiks\_cube} & Can, Foam Brick, Mug, Apple, Ball, Orange \\
        \textbf{Can} & \texttt{010\_potted\_meat\_can} & Foam Brick, Mug, Apple, Ball, Orange, Pear \\
        \textbf{Pear} & \texttt{016\_pear} & Can, Foam Brick, Mug, Apple, Ball, Cube \\
        \textbf{Mug} & \texttt{025\_mug} & Can, Foam Brick, Apple, Ball, Orange, Cube \\
        \textbf{Lego} & \texttt{073-e\_lego\_duplo} & Fish Can, Foam Brick, Mug, Apple, Ball, Orange \\
        \textbf{Ball} & \texttt{053\_mini\_soccer\_ball} & Can, Foam Brick, Mug, Apple, Pear, Cube \\
        \bottomrule
    \end{tabular}
\end{table}

\subsubsection{Simulation Parameters}
The benchmark is built on ManiSkill 3 (SAPIEN engine) with the following physical parameters to mimic real-world friction and contact dynamics:
\begin{itemize}
    \item \textbf{Target Friction:} Static $\mu_s=2.0$, Dynamic $\mu_d=2.0$ (High friction to facilitate grasping).
    \item \textbf{Clutter Friction:} Static $\mu_s=1.0$, Dynamic $\mu_d=1.0$ (Moderate friction to allow pushing/sliding).
    \item \textbf{Restitution:} 0.0 for all objects (inelastic collisions).
    \item \textbf{Damping:} Linear damping 0.1, Angular damping 0.1.
\end{itemize}